\documentclass[11pt,a4paper]{article}
\usepackage[hyperref]{emnlp2020}
\usepackage{times}
\usepackage{latexsym}

\usepackage{multirow}
\usepackage{multicol}
\usepackage{caption,subcaption}
\usepackage{calrsfs}
\usepackage{mathrsfs}
\usepackage{graphicx}
\usepackage{algorithm}
\usepackage{algorithmic}
\usepackage[symbol]{footmisc}
\usepackage{url}            % simple URL typesetting
\usepackage{booktabs}       % professional-quality tables
\usepackage{nicefrac}       % compact symbols for 1/2, etc.
\usepackage{microtype}      % microtypography
\usepackage{amsmath,amssymb,amsfonts}
\usepackage{amsthm}
\usepackage{xspace}
\usepackage{xcolor}
\DeclareMathAlphabet{\pazocal}{OMS}{zplm}{m}{n}
\newcommand{\Lb}{\pazocal{L}}   % different
   
\usepackage{amsmath,amsfonts,bm}

\newcommand{\RNum}[1]{\uppercase\expandafter{\romannumeral #1\relax}}

\newcommand{\xhdr}[1]{{\noindent\bfseries #1}.}
\newcommand{\cut}[1]{}

\newcommand{\removelatexerror}{\let\@latex@error\@gobble}

\usepackage{colortbl}
\definecolor{cobalt}{rgb}{0.0, 0.28, 0.67}

\definecolor{mistyrose}{rgb}{1, 0.92, 0.92}

\definecolor{grannysmithapple}{rgb}{0.79, 0.92, 0.77}

\usepackage{amsmath,amsfonts,bm}
\newcommand\restr[2]{{% we make the whole thing an ordinary symbol
  \left.\kern-\nulldelimiterspace % automatically resize the bar with \right
  #1 % the function
  \vphantom{\big|} % pretend it's a little taller at normal size
  \right|_{#2} % this is the delimiter
  }}
\def\1{\bm{1}}

\DeclareMathAlphabet{\mathsfit}{\encodingdefault}{\sfdefault}{m}{sl}
\SetMathAlphabet{\mathsfit}{bold}{\encodingdefault}{\sfdefault}{bx}{n}

\usepackage{microtype}
\aclfinalcopy % Uncomment this line for the final submission
\title{Structure Aware Negative Sampling in Knowledge Graphs}

\author{Kian Ahrabian\thanks{$\text{\;}$ Equal contribution, names ordered alphabetically.}$^{\text{\; },1,3}$, Aarash Feizi\footnotemark[1]$^{\text{\; },1,3}$, Yasmin Salehi\footnotemark[1]$^{\text{\; },2}$, \\ {\bf William L. Hamilton$^{1,3,4}$} \and {\bf Avishek Joey Bose$^{1,3}$} \\ 
$^1$ School of Computer Science, McGill University, Canada \\
$^2$ Department of Electrical and Computer Engineering, McGill University, Canada \\
$^3$ Montreal Institute of Learning Algorithms (Mila), Canada \\
$^4$ Canada CIFAR AI Chair \\
\{\texttt{kian.ahrabian}, \texttt{aarash.feizi}, \texttt{yasmin.salehi} \} \\
\{\texttt{@mail.mcgill.ca} \}
}

\date{}

\begin{document}
\maketitle
\begin{abstract}
Learning low-dimensional representations for entities and relations in knowledge graphs using contrastive estimation represents a scalable and effective method for inferring connectivity patterns. A crucial aspect of contrastive learning approaches is the choice of corruption distribution that generates hard negative samples, which force the embedding model to learn discriminative representations and find critical characteristics of observed data. While earlier methods either employ too simple corruption distributions, i.e. uniform, yielding easy uninformative negatives or sophisticated adversarial distributions with challenging optimization schemes, they do not explicitly incorporate known graph structure resulting in suboptimal negatives. In this paper, we propose Structure Aware Negative Sampling (SANS), an inexpensive negative sampling strategy that utilizes the rich graph structure by selecting negative samples from a node's $k$-hop neighborhood. Empirically, we demonstrate that SANS finds semantically meaningful negatives and is competitive with SOTA approaches while requires no additional parameters nor difficult adversarial optimization.

\end{abstract}

\section{Introduction}

Knowledge Graphs (KGs) are repositories of information organized as factual triples ($h,r,t$), where head and tail entities are connected via a particular relation ($r$). Indeed, KGs have seen wide application in a variety of domains such as question answering \cite{yao2014information, hao2017end, moldovan2001logic} and machine reading \cite{weissenborn2018jack, yang2017leveraging} to name a few and have a rich history within the natural language processing (NLP) community \cite{berant2013semantic, yu2014improving, collobert2008unified, peters2019knowledge}. While often large, real-world KGs such as FreeBase \cite{bollacker2008freebase} and WordNet \cite{miller1995wordnet} are known to be incomplete. Consequently, KG completion via link prediction constitutes a fundamental research topic ameliorating the practice of important NLP tasks \cite{sun2019rotate, angeli2013philosophers}.

% old version start
\cut{Indeed, KGs have seen wide application in a variety of domains, such as question answering \cite{yao2014information, hao2017end}, and natural language processing \cite{berant2013semantic, yu2014improving}, to name a few. While often large, real-world KGs such as FreeBase \cite{bollacker2008freebase} and WordNet \cite{miller1995wordnet} are known to be incomplete. Consequently, KG completion via link prediction constitutes a fundamental task \cite{sun2019rotate, kotnis2017analysis, angeli2013philosophers}.}
% old version end

In recent years, there has been a surge of methods employing graph embedding techniques that encode KGs into a lower-dimensional vector space facilitating easier data manipulation \cite{zhang2019nscaching} while being an attractive framework for handling data sparsity and incompleteness \cite{wang2018incorporating}. To learn such embeddings, contrastive learning has emerged as the de facto gold standard. Indeed, contrastive learning approaches enjoy significant computational benefits over methods that require computing an exact softmax over a large candidate set, such as over all possible tail entities given a head and relation. Another important consideration is modeling needs, as certain assumptions are best expressed as some score or energy in margin-based or un-normalized probability models \cite{smith2005contrastive}. For example, modeling entity relations as translations or rotations in a vector space naturally leads to a distance-based score to be minimized for observed entity-relation-entity triplets \cite{bordes2013translating}.

Leveraging contrastive estimation to train KG embedding models involves optimizing the model by pushing up the energy with respect to observed positive triplets while simultaneously pushing down energy on negative triplets. Consequently, the choice of negative sampling distribution plays a crucial role in shaping the energy landscape as simple random sampling---e.g. Noise Contrastive Estimation (NCE) \cite{gutmann2010noise}---produces negatives that are easily classified and provide little information alongside in the form of a gradient signal. This is easily remedied if the corruption process selects a hard negative example through more complex negative sampling distribution, such as adversarial samplers \cite{cai2017kbgan, bose2018adversarial, sun2019rotate}.\cut{or even refining the sampling process by using cached high-quality negative triplets \cite{zhang2019nscaching}.} However, adversarial negative sampling methods are computationally expensive, while more tractable approaches---e.g. cache-based methods \cite{zhang2019nscaching}---are not tailored to the KG setting as they fail to incorporate known graph structure as part of the sampling process. This raises the important question of whether we can obtain a computationally inexpensive negative sampling strategy while benefiting from the rich graph structure of KGs.

\xhdr{Present Work}
In this work, we introduce \emph{Structure Aware Negative Sampling} (SANS), an algorithm that utilizes the graph structure of a KG to find hard negative examples. Specifically, SANS constructs negative samples using a subset of entities restricted to either the head or tail entity's \emph{k}-hop neighborhood. We hypothesize that entities that are within each other's neighborhood but share no direct relation have higher chances of being related to one another and thus are good candidates for negative sampling. We also experiment with a dynamic sampling scheme based on random walks to approximate a node's local neighborhood. Empirically, we find that negative sampling using SANS consistently leads to improvements upon uniform sampling and sophisticated Generative Adversarial Network \cite{goodfellow2014generative} (GAN) based approaches at a fraction of the computational cost, and is competitive with other SOTA approaches with no added parameters.

\begin{figure}[t]
    \centering
    \includegraphics[width=3in]{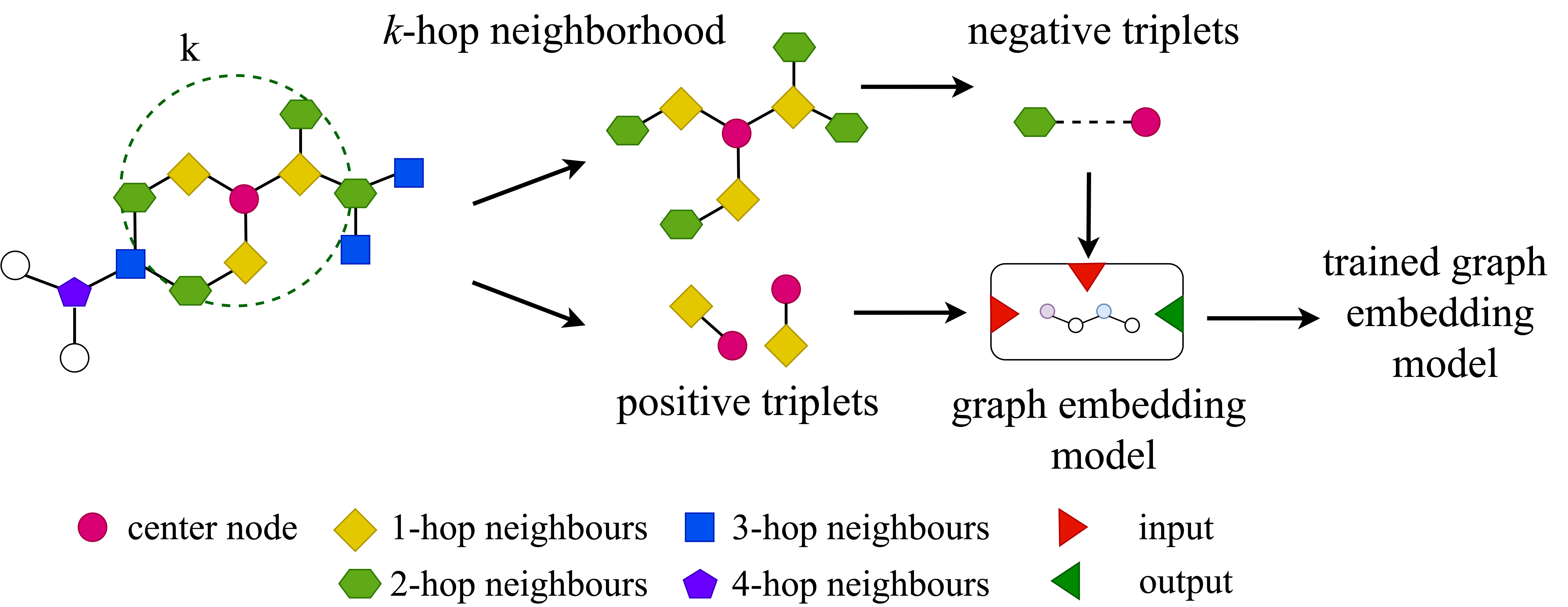}
    \caption{Our proposed approach for training a graph embedding model. In this illustration, $k$ is set to $2$.}
    \label{fig:arch}
\end{figure}

\cut{
Inspired by GAN-based and cache-based approaches, which all aim to select high-quality negative triplets, in this work, the effect of generating negative samples with respect to nodes' neighborhood is investigated, hence why our proposed approach is called \emph{Structural Negative Sampling}. The heuristic used while performing sampling is that nodes that are within each other's neighborhood have higher chances of being related to one another. In this regard, two nodes that are in the same neighborhood of an arbitrary distance $r$ that are not connected can make a negative triplet that has a high score (evaluated from the embedding model score functions) and is of high-quality. Therefore, not only our proposed negative sampling scheme benefits from using the underlying graph structure, it is also simple and inexpensive to train yet effective. This idea is analogous to how negative triplets are generated in DeepWalk \cite{perozzi2014deepwalk} and node2vec \cite{grover2016node2vec}, in which the node degree or word frequency is considered respectively. A high-level architectural diagram of our proposed negative sampling scheme is shown in Figure \ref{fig:arch}.
}

\section{Related Work}
\label{sec:background}
\xhdr{Negative Sampling} 
Negative sampling is a method that can be employed to enable the scaling of log-linear models. In essence, negative sampling resolves computational intractability of computing the normalization constant by changing the task to distinguishing observed positive data and fictitious negative examples that are generated by corrupting the positive examples. This general approach is a simplification of NCE, which is based on a Monte-Carlo approximation of the partition function used in Importance Sampling (IS) \cite{bengio2003quick}.

% ----------------------------------------------------------------------------------------------------------------------------------------------------------------------------------
\xhdr{Non-Fixed Negative Sampling}
\label{sec:relatedwork}
As proposed in \cite{mikolov2013distributed}, negative triplets can be generated using a uniform sampling scheme. However, such uniform and fixed sampling schemes result in easily-classified negative triplets during training, which do not provide any meaningful information \cite{sun2019rotate, zhang2019nscaching}. Hence, as the training progresses, most of the sampled negative triplets receive small scores and almost zero gradients, impeding the training of the graph embedding model after only a small number of iterations. 

To address the issue of easy negatives, ~\citet{sun2019rotate} propose Self-Adversarial negative sampling, which weighs each sampled negative according to its probability under the embedding model. Alternatively, the authors in \cite{wang2018incorporating} and \cite{cai2017kbgan} try creating high-quality negative samples by exploiting GANs, which, while effective, are expensive to train and require black-box gradient estimation techniques. Another elegant approach that uses fewer parameters and is easier to train compared to GAN-based methods is NSCaching \cite{zhang2019nscaching}, which involves using a cache of high-quality negative triplets---i.e. those with high scores.

\section{Structure Aware Negative Sampling}
\label{sec:proposedapproach}
\cut{Given an observed positive triplet $(h,r,t)$, a negative sample can be constructed by corrupting either the head or tail entity to form a new triplet---i.e. $(h',r,t')$ as such:
\begin{equation}
    (h',r,t') = \{ (h',r,t)|h'\in E\} \cup \{(h,r,t')|t'\in E \}  \\
\end{equation}
where $E$ is the set of all entities in the KG.
}
Given an observed positive triplet $(h,r,t)$, a negative sample can be constructed by corrupting either the head or tail entity to form a new triplet---i.e. $(h',r,t')$---where either $h',t' \in E$, where $E$ is the set of all entities in the KG.
\cut{
\begin{equation}
    (h',r,t') = \{ (h',r,t)|h'\in E\} \cup \{(h,r,t')|t'\in E \}  \\
\end{equation}
}
Additionally, we assume that the graph embedding models are trained using a loss function of the following form:
\begin{align}
    % \Lb &= -\sum_{(h,r,t) \in S} \text{log }\sigma(\gamma - d_{r}(\boldsymbol{h},\boldsymbol{t}))  \notag \\ 
    \Lb &= - \ \text{log }\sigma(\gamma - d_{r}(\boldsymbol{h},\boldsymbol{t}))  \notag \\ 
    &- \sum_{i=1}^{n} \frac{1}{n}\text{log }\sigma(d_r(\boldsymbol{h}_i^{'},\boldsymbol{t}_i^{'})-\gamma)
\end{align}
where $d_r(h,t)$ denotes the score assigned to the compatibility of head and tail entities under the relation $r$, $\gamma$ is a fixed margin, $\sigma$ is the sigmoid function, and $n$ is the number of negative samples.

In this paper, we seek to explicitly use the rich graph structure surrounding a particular node when generating negative triplets. We motivate our approach based on the observation that prior work in learning word embeddings \cite{mikolov2013distributed}, where negative sampling has historically developed, lacked the richness of graph structure that is immediately accessible in the KG setting. Consequently, we hypothesize that enriching the negative sampling process with structural information can yield harder negative examples, crucial to learning effective embeddings. Fig. \ref{fig:arch} highlights our approach, which requires the construction of the $k$-hop neighborhood ($\mathbb{K}$) for each node at its first step,
\begin{equation}
\label{eqn:khop}
\mathbb{K} = S^+(A^{k} + A^{k-1})
\end{equation}
for $k>0$, where $k$ is an integer, representing the neighborhood radius, $A$ is the KG's adjacency matrix, and $S^+$ is the element-wise sign function set to $1$ if a path exists and $0$ otherwise.

To construct negatives triplets, we may now simply sample from the nonzero cell of $\mathbb{K}$, which represents a subset of all entities for each node in the KG---i.e. $\mathbb{K} \subset \mathbf{1}^{E \times E}$.\cut{Thus, selected negative triples satisfy the following criterion:
\begin{equation*}
    (h',r,t') = \{ (h',r,t)|h'\in \mathbb{K}\} \cup \{(h,r,t')|t'\in \mathbb{K} \}  \\
\end{equation*}
}
Intuitively, SANS exploits the locality of an entity's neighborhood, where negative samples are defined as entities that are not directly linked under a relation $r$ but can be accessed through a path of at most length $k$. We argue that such local negatives are harder to distinguish and lead to higher scores as evaluated by the embedding model. One important technical detail in constructing $\mathbb{K}$ is the existence of multiple relation types, which requires an additional dimension to represent the graph connectivity as adjacency and \emph{k}-hop tensors.

\subsection{Variants of SANS}

% old version start
\cut{Although SANS requires a one-time preprocessing step to construct $\mathbb{K}$ as defined in Eqn. \ref{eqn:khop}, this may still be costly for large and dense KGs. To combat this inefficiency, we introduce \emph{RW-SANS}, which uses random walks of length $k$ in the adjacency tensor \cite{perozzi2014deepwalk} to approximate the \emph{k}-hop neighborhood.\cut{ Thus, the number of random walks trades off the computational cost with approximation accuracy of the true $\mathbb{K}$.}}
% old version end

Although SANS requires a one-time preprocessing step to construct $\mathbb{K}$ as defined in Eqn. \ref{eqn:khop}, this may still be costly for large and dense KGs. To combat this inefficiency, we introduce \emph{RW-SANS} in Alg. \ref{alg:rw}, which uses $\omega$ random walks \cite{perozzi2014deepwalk} of length $k$ in the adjacency tensor to approximate the \emph{k}-hop neighborhood.\cut{ Thus, the number of random walks trades off the computational cost with approximation accuracy of the true $\mathbb{K}$.}
 
\begin{algorithm}[ht]
    \caption{Approximating the \emph{k}-hop Neighborhood Using Random Walks}
    \begin{algorithmic}
    \STATE {\bfseries Input:} $A, R, k, \omega$
    \COMMENT A: adjacency tensor, R: set of relation types, k: \# of \emph{k}-hops, $\omega$: \# of random walks
    \STATE {$K \gets$ sparseTensor($|A| \times|R| \times |A|$)}
    \FORALL {entity $e$}
    \STATE {$K[e] \gets $} randomWalk(k, $\omega$)
    \ENDFOR
    \STATE {\textbf{return}} $K$
    \end{algorithmic}
\label{alg:rw}
\end{algorithm}

As SANS constructs a local neighborhood from which negative samples are drawn, it can also be combined with other negative sampling approaches. In this work, we extend the Self-Adversarial approach in \cite{sun2019rotate} and combine it with SANS by restricting the negative triplet candidate set to the \emph{k}-hop neighborhood. In the subsequent sections, we refer to this technique as \emph{Self-Adversarial (Self-Adv.) SANS}, whereas the former approach is referred to as \emph{Uniform SANS}. 

\cut{
Computing Eq. \ref{eqn:khop} can be costly for large KGs, the \emph{k-hop} neighborhood can be approximated using \emph{Random Walks (RW)} \cite{perozzi2014deepwalk} by Algorithm \ref{alg:rw}, where $\omega$ is the number of RWs, and $k$ is the number of k-hops.

\begin{algorithm}[tb]
   \caption{\small Approximating K-hop neighborhood using RWs}
   \label{alg:rw}
  \begin{small}
\begin{algorithmic}
   \STATE {\bfseries Input:} $A, k, \omega$
   \COMMENT A: Adjacency Matrix, k: \# of k-hops, $\omega$: \# of RWs
   \STATE {$K \gets$ sparseMatrix($|A| \times |A|$)}
   \FORALL {entity $e$}
   \STATE {$K[e] \gets $} randomWalk(k, $\omega$)
   \ENDFOR
   \STATE {\textbf{return}} $K$
\end{algorithmic}
\end{small}
\end{algorithm}}

\section{Experiments}
\label{sec:results}

% MAIN RESULT TABLE 1
\begin{table*}[ht]
\begin{small}
\centering
\begin{tabular}{cccccccc}
\hline
\multirow{3}{2cm}{\centering Score Function}& \multirow{3}{2cm}{Algorithm} & \multicolumn{2}{c}{FB15K-237} & \multicolumn{2}{c}{WN18} & \multicolumn{2}{c}{WN18RR} \\
& & Hit@10 & \multirow{2}{1cm}{\centering MRR} & Hit@10 & \multirow{2}{1cm}{\centering MRR} & Hit@10 & \multirow{2}{1cm}{\centering MRR} \\ 
& & (\%) & & (\%) & & (\%) & \\
\hline
% TransE --------------------------------------------------
\multirow{6}{1.5cm}{\centering TransE} 
& KBGAN \cite{cai2017kbgan} & 46.59 & 0.2926 & 94.80 & 0.6606 & 43.24 & 0.1808 \\
& NSCaching \cite{zhang2019nscaching} & 47.64 & \textbf{0.2993} & 94.63 & \textbf{0.7818} & 47.83 & 0.2002\\
& Uniform \cite{sun2019rotate} & \textbf{48.03} & 0.2927 & \textbf{95.53} & 0.6085 & \textbf{49.63 }& \textbf{0.2022} \\
& Uniform SANS (ours) & 48.35 & 0.2962 & 95.09 & \textbf{0.8228$^\star$} & 51.15 & 0.2254 \\
& Uniform RW-SANS (ours) & \textbf{48.50$^\star$}& \textbf{0.2981$^\star$} & \textbf{95.22$^\star$} &	0.8195 &	\textbf{53.41$^\star$} & \textbf{0.2317$^\star$}  \\
& $\Delta$ & \cellcolor{grannysmithapple}+0.47& \cellcolor{mistyrose}-0.0012& \cellcolor{mistyrose}-0.31 & \cellcolor{grannysmithapple}+0.0410 & \cellcolor{grannysmithapple}+3.78 & \cellcolor{grannysmithapple}+0.0295 \\
\hline
% DistMult --------------------------------------------------
\multirow{6}{1.5cm}{\centering DistMult} & 
KBGAN & 39.91 & 0.2272 & 93.08 & 0.7275  & 29.52 & 0.2039 \\
& NSCaching & \textbf{45.56} & \textbf{0.2834} & \textbf{93.74} & \textbf{0.8306} & 45.45 & \textbf{0.4128} \\
& Uniform & 40.26 & 0.2537 & 81.39  & 0.4689 & \textbf{52.86} & 0.3938 \\
& Uniform SANS (ours) & 41.00 & 0.2595 & \textbf{93.19$^\star$} & \textbf{0.7553$^\star$} & 44.74 & 0.4025 \\
& Uniform RW-SANS (ours) & \textbf{41.46$^\star$} &	\textbf{0.2621$^\star$} &	89.80 &	0.6235	& \textbf{49.09$^\star$}	& \textbf{0.4071$^\star$}  \\
& $\Delta$ & \cellcolor{mistyrose}-4.10 & \cellcolor{mistyrose}-0.0213 & \cellcolor{mistyrose}-0.55 & \cellcolor{mistyrose}-0.0753 & \cellcolor{mistyrose}-3.77 & \cellcolor{mistyrose}-0.0057 \\
\hline
% RotatE --------------------------------------------------
\multirow{4}{1cm}{\centering RotatE} & Uniform & \textbf{47.85} & \textbf{0.2946} & \textbf{96.09} & \textbf{0.9474} & \textbf{56.51} & \textbf{0.4711} \\
& Uniform SANS (ours) & 48.22& 0.2985 & 95.97 & \textbf{0.9499$^\star$} & 55.76 & 0.4769 \\
& Uniform RW-SANS (ours) & \textbf{48.47$^\star$}	& \textbf{0.3003$^\star$} &	\textbf{96.07$^\star$} &	0.9489 &	\textbf{57.12$^\star$}	& \textbf{0.4796$^\star$}  \\
& $\Delta$ & \cellcolor{grannysmithapple}+0.62 &	\cellcolor{grannysmithapple}+0.0057 & \cellcolor{mistyrose}-0.02 & \cellcolor{grannysmithapple}+0.0025 & \cellcolor{grannysmithapple}+0.61 & \cellcolor{grannysmithapple}+0.0085 \\
\hline
\end{tabular}
\caption{Comparison of different negative sampling algorithms. \textbf{Bold}  and \textbf{marked bold$^\star$} numbers represent the best  SOTA and SANS algorithms respectively.  % \vspace{4mm} Results for KBGAN and NSCaching are the \textit{scratch} values taken from \cite{zhang2019nscaching}.
}
\label{tab:unisota_comp}
\end{small}
\end{table*}
% ------------------------------------------------------------

% RESULTS TABLE 2 START
% MAIN RESULT TABLE
\begin{table*}[ht]
\begin{small}
\centering
\begin{tabular}{cccccccc}
\hline
\multirow{3}{2cm}{\centering Score Function}& \multirow{3}{2cm}{Algorithm} & \multicolumn{2}{c}{FB15K-237} & \multicolumn{2}{c}{WN18} & \multicolumn{2}{c}{WN18RR} \\
& & Hit@10 & \multirow{2}{1cm}{\centering MRR} & Hit@10 & \multirow{2}{1cm}{\centering MRR} & Hit@10 & \multirow{2}{1cm}{\centering MRR} \\ 
& & (\%) & & (\%) & & (\%) & \\
\hline
% TransE --------------------------------------------------
\multirow{4}{1.5cm}{\centering TransE} &
Self-Adv. \cite{sun2019rotate} & \textbf{52.73} & \textbf{0.3296} & \textbf{92.02} & \textbf{0.7722} & \textbf{52.78} & \textbf{0.2232} \\
& Self-Adv. SANS (ours) & \textbf{52.03$^\star$} & \textbf{0.3265$^\star$} & 84.06 & 0.7136 & 53.21 & 0.2249 \\
& Self-Adv. RW-SANS (ours) & 50.04 & 0.3060 &\textbf{88.51$^\star$}&	\textbf{0.7429$^\star$}&	\textbf{53.81$^\star$}&\textbf{0.2273$^\star$} \\
& $\Delta$ & \cellcolor{mistyrose}-0.70 & \cellcolor{mistyrose}-0.0031 &	\cellcolor{mistyrose}-3.51 &	\cellcolor{mistyrose}-0.0293 & \cellcolor{grannysmithapple}+1.03 & \cellcolor{grannysmithapple}+0.0041 \\
\hline
% DistMult --------------------------------------------------
\multirow{4}{1.5cm}{\centering DistMult} & Self-Adv. & \textbf{48.41} & \textbf{0.3091} & \textbf{92.94}  & \textbf{0.6837} & \textbf{53.80} & \textbf{0.4399} \\
& Self-Adv. SANS (ours) & \textbf{48.68$^\star$} & \textbf{0.3100$^\star$} & \textbf{93.04$^\star$} & \textbf{0.7561$^\star$} & 38.70 & 0.3684 \\
& Self-Adv. RW-SANS (ours) & 48.17 &	0.3071 &	91.08 &	0.6634	& \textbf{42.74$^\star$}	&\textbf{0.3836$^\star$}\\
& $\Delta$ & \cellcolor{grannysmithapple}+0.27 &\cellcolor{grannysmithapple}+0.0009 & \cellcolor{grannysmithapple}+0.10 & \cellcolor{grannysmithapple}+0.0724 & \cellcolor{mistyrose}-11.06 & \cellcolor{mistyrose}-0.0563 \\
\hline
% RotatE --------------------------------------------------
\multirow{4}{1cm}{\centering RotatE} & Self-Adv. & \textbf{53.03} & \textbf{0.3362} & \textbf{96.05}& \textbf{0.9498} & \textbf{57.29} & \textbf{0.4760} \\
& Self-Adv. SANS (ours) & \textbf{53.12$^\star$} & \textbf{0.3358$^\star$} & 95.85 & 0.9494 & \textbf{57.12$^\star$}& 0.4745 \\
& Self-Adv. RW-SANS (ours) & 51.07 &	0.3161	&\textbf{96.09$^\star$}&	\textbf{0.9496$^\star$}&	56.94&	\textbf{0.4805$^\star$} \\
& $\Delta$ & \cellcolor{grannysmithapple}+0.09 &\cellcolor{mistyrose} -0.0004	& \cellcolor{grannysmithapple}+0.04 &\cellcolor{mistyrose}-0.0002 & \cellcolor{mistyrose}-0.17 & \cellcolor{grannysmithapple}+0.0045\\
\hline
\end{tabular}
\caption{Comparison of the Self-Adversarial negative sampling technique with our Self-Adversarial SANS. \textbf{Marked bold$^\star$} numbers  are the results of the best SANS implementation. 
}
\label{tab:comp_adv}
\end{small}
\end{table*}
% RESULTS TABLE 2 END

We investigate the application of SANS-based negatives to train KG embedding models based on the TransE, DistMult, and RotatE models for the task of KG completion\footnote{Code available at https://github.com/kahrabian/SANS}. We evaluate our proposed approach on standard benchmarks, consisting of FB15K-237 \cite{bollacker2008freebase}, WN18 and WN18RR \cite{miller1995wordnet}. From our experiments we seek to answer the following questions: 
\begin{itemize}
    \item[\textbf{(Q1)}] \textbf{Hard Negatives: Can we sample hard negatives purely using graph structure?}
    \item[\textbf{(Q2)}] \textbf{Can we combine graph structure with other SOTA negative samplers?} 
    \item[\textbf{(Q3)}] \textbf{Can we effectively approximate the adjacency tensor with random walks?}
\end{itemize}
In our experiments, we rely on three representative baselines, namely uniform negative sampling \cite{bordes2013translating}, KBGAN \cite{cai2017kbgan}, and NSCaching \cite{zhang2019nscaching}. We also compare with the current SOTA approach in Self-Adversarial negative sampling \cite{sun2019rotate}, and we test whether local graph structure can also be leveraged in this setting. \cut{As evaluation measures we use Hits at N (H@N), and Mean Reciprocal Rank (MRR).}

\subsection{Results}
We now address the core experimental questions.

% Old version start
\cut{
\xhdr{Q1:} Table \ref{tab:unisota_comp} summarizes our main results where we highlight SANS and RW-SANS. We also compute the difference between the best variant of SANS against the best performing baseline in row $\Delta$. Overall, we find that SANS negatives almost always lead to harder negative samples over Uniform and KBGAN negatives on all three datasets. Furthermore, SANS achieves competitive performance with NSCaching without any new parameters. Lastly, we observe an average $\Delta$ value of $0.0231$, $-0.0341$, and $0.0056$ in MRR for TransE, DisMult, and RotateE respectively, thus confirming our hypothesis that graph structure indeed helps.}
% Old versionn end 

\xhdr{Q1:} Table \ref{tab:unisota_comp} summarizes our main quantitative results where we highlight SANS and RW-SANS. We also compute the difference between the best variant of SANS against the best performing baseline in row $\Delta$. Overall, we find that SANS negatives almost always lead to harder negative samples over Uniform and KBGAN negatives on all three datasets. Furthermore, SANS achieves competitive performance with NSCaching when combined with TransE, and is the second best-performing algorithm when combined with DistMult without requiring additional parameters. We observe average $\Delta$ values of $0.0231$, $-0.0341$, and $0.0056$ in MRR for TransE, DisMult, and RotateE respectively, which confirm our approach's effectiveness compared to SOTA while remaining computationally efficient.   

We also qualitatively investigate the semantic hardness of SANS negatives against negatives generated via uniform sampling. For instance, using the center node ``arachnoid" in the WN18RR dataset as an example, the negatives sampled via SANS within a 2-hop neighborhood are “arachnida,” “biology,” “arthropod,” “wolf spider,” and “garden spider,” while the ones picked by uniform sampling are “diner,” “refusal,” “landscape,” “rise,” and “nurser.” Clearly, the negatives found via SANS are semantically harder to distinguish, and as a result, they also confirm the importance of incorporating graph structure into negative samplers to aid in `hard' negative mining. A more detailed qualitative analysis of negative samples---including the effect of varying neighborhood sizes---generated by SANS can be found in \ref{qualitative_sans}.

% Old version start ------------
\cut{
\xhdr{Q2:} We now combine our approach SANS with Self-Adversarial negative sampling \cite{sun2019rotate}. Our results are presented in Table \ref{tab:comp_adv} under Self-Adv. SANS and Self-Adv. RW-SANS, both of which reweigh the negative triplets as done in \cite{sun2019rotate}. \cut{under embedding models score.} We observe comparable performance between the two approaches, but crucially this is achieved by mostly considering $0.2\%$ to $9\%$ of the entities in the datasets like in WN18 and WN18RR (see Table \ref{tab:percentage} in Appendix), further highlighting the importance of utilizing graph structure.
}
% Old version end ------------

\xhdr{Q2:} We now combine our approach SANS with Self-Adversarial negative sampling \cite{sun2019rotate}. Our results are presented in Table \ref{tab:comp_adv} under Self-Adv. SANS and Self-Adv. RW-SANS, both of which reweigh the negative triplets as done in \cite{sun2019rotate}. \cut{under embedding models score.} We observe comparable performance between the two approaches, but crucially this is achieved by mostly considering $0.2\%$ to $9\%$ of the entities in the datasets like in WN18 and WN18RR, as indicated in Table \ref{tab:percentage}. By considering that the partially-filled adjacency tensors improve computational feasibility for requiring less memory and allowing sparse tensor operations to take place, the appeal of incorporating graph structure while choosing negative samples is further highlighted.

% % % ------------------------------------------------------------------------
% TABLE START
% % of filled entries
\begin{table}[htbp]
\centering
\begin{tabular}{cccccc}
\hline
$k$ & 2 & 3 & 4 & 5 \\ 
\hline
FB15K-237 & 34 & 83 & 97 & 99 \\
\hline
WN18 & 0.19 & 0.75 & 3.22 & 10.2 \\
\hline
WN18RR & 0.16 & 0.65 & 2.76 & 8.67 \\
\hline
\end{tabular}
\caption{Percentage (\%) of filled entries in the \emph{k}-hop adjacency tensor.}
\label{tab:percentage}
\end{table}
% TABLE END
% % % ------------------------------------------------------------------------

\begin{figure}[h]
    \centering
    \includegraphics[width=2.5in]{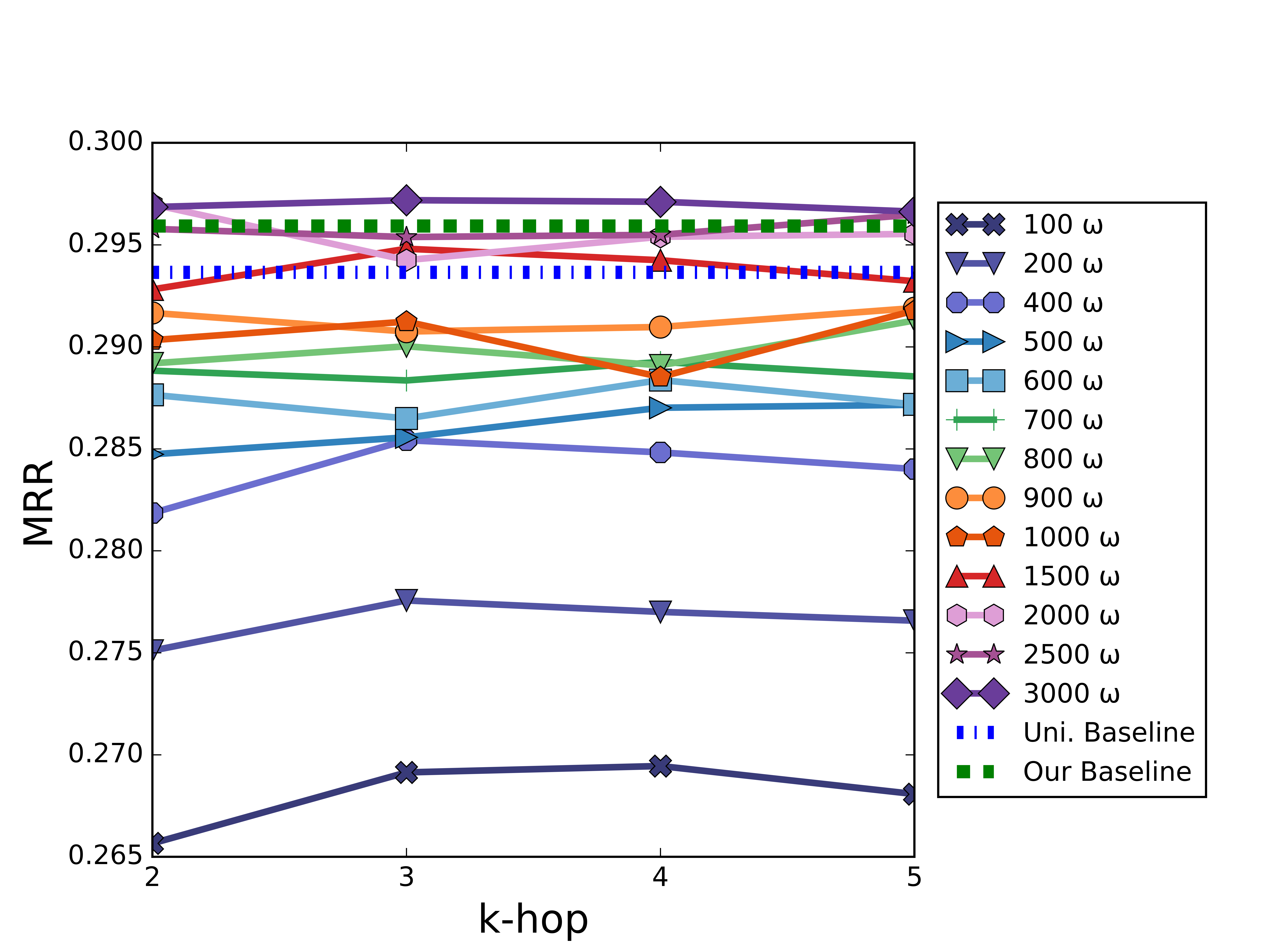}
    \caption{The performance of Uniform RW-SANS with TransE on FB15K-237 using different $\omega$ values.
    \cut{The effect of the number of random walks
     for Uniform SANS with TransE on FB15K-237.}}
    \label{fig:ablation}
\end{figure}

\xhdr{Q3:} We now analyze the impact of approximating the local neighborhood using random walks. Fig. \ref{fig:ablation} depicts the effect of varying the number of random walks ($\omega$) with neighborhoods of different radii and MRR. We report two baselines, one being the performance of uniform sampling, and the other being our best performance achieved by Uniform SANS when combined with TransE, for which the \emph{k}-hop tensor was explicitly computed. Interestingly, we find that the \emph{k}-hop tensor can not only be well approximated with 3000 random walks, but RW-SANS beats both baselines. We reconcile this result by noting that certain nodes have a higher probability of being sampled due to sharing a larger number of paths with the center node, resulting in an implicit weighted negative sampling scheme.

\section{Conclusion and Future Directions}
\label{sec:conclusion}

In this work, we introduced SANS, a novel negative sampling strategy, which directly leverages information about \emph{k}-hop neighborhoods to select negative examples. Our work sheds light on the need and importance of incorporating graph structure when designing negative samplers for KGs, and for which SANS can be seen as a cheap yet powerful baseline that requires no additional parameters or difficult optimization. Empirically, we find that SANS-based negatives have comparable performance with SOTA approaches and even outperform previous sophisticated GAN-based approaches.

\section*{Acknowledgments}

The authors would like to thank the anonymous EMNLP reviewers for their helpful and constructive feedback. This research was supported by a Canada CIFAR AI Chair and NSERC Discovery Grant RGPIN-2019-0512. Avishek Joey Bose is also generously supported through the IVADO PhD fellowship.

% The acknowledgments should go immediately before the references. Do not number the acknowledgments section.
% Do not include this section when submitting your paper for review.

% \bibliography{anthology,emnlp2020}
\bibliography{emnlp2020}
\bibliographystyle{acl_natbib}

% -------------------------------------------------------
% \newpage 

\appendix

% \onecolumn

% \newpage
%% ******** NOTE **********
%% t_xxx.tex files are the tables. 
\appendix
%\section{Supplemental Material}
\label{sec:supplemental}

\section{Experimental Settings}

This section provides an overview of the datasets and evaluation protocols used for obtaining our results. 

\subsection{Datasets}

To conduct experiments for our proposed methods, datasets FB15K-237, WN18, and WN18RR were used. FB15K-237 is a subset of FB15K, which has been derived from the FreeBase Knowledge Base (KB) \cite{bollacker2008freebase}, a large database that contains general facts about the world with many different relation types. On the other hand, WN18RR is a subset of WN18, which has been derived from the WordNet KB \cite{miller1995wordnet}, which is a large lexical English database that captures lexical relations---e.g. the super-subordinate relations between words. The WN18 and FB15K were first introduced in \cite{bordes2013translating} and were used in the majority of KG-related researches. In comparison, WN18 and WN18RR contain less relation types than FB15K-237. A summary of the number of entities and relation types corresponding to each of these datasets is provided in Table \ref{tab:datasets}.

\begin{table}[H]
\centering
\begin{tabular}{cccccc}
\hline
Dataset   & \#entity & \#relation \\ \hline
FB15K-237 & 14,541 & 237  \\
WN18 & 40,943 & 18 \\
WN18RR & 40,943 & 11  \\
\hline 
\end{tabular}
\caption{Dataset Information \cite{sun2019rotate}.}
\label{tab:datasets}
\end{table}

 % the dataset table

\subsection{Evaluation Protocols}
To evaluate our negative sampling approach, we used standard evaluation metrics, consisting of Mean Reciprocal Rank (MRR) and Hits at N (H@N). The train/validation/test split information is provided in Table \ref{tab:train_val_test_split}.

\begin{table}[H]
\centering
\begin{tabular}{cccccc}
\hline
Dataset   & \#training & \#validation & \#test \\ \hline
FB15K-237 & 272,115  & 17,535  & 20,466 \\
WN18 & 141,442 & 5,000 & 5,000 \\
WN18RR & 86,835 & 3,034 & 3,134 \\
\hline 
\end{tabular}
\caption{Train/Validation/Test Split Information \cite{sun2019rotate}.}
\label{tab:train_val_test_split}
\end{table} % The train, validation, test split table

% ---------------------------------------------------------

\section{Implementation Details}

This section of the supplemental goes over the implementation details of our RW-SANS algorithms---i.e. Uniform RW-SANS and Self-Adv. RW-SANS, which use random walks to approximate the \emph{k}-hop adjacency tensor. Other experimental setups are further detailed herein.

% ----------------------------------------------------------

\subsection{Hyperparameters}

$k$ and $\omega$ (for when the \emph{k}-hop neighborhood is being approximated by Alg. \ref{alg:rw}) are the hyperparameters in our negative sampling algorithms. To find the optimal hyperparameter values that resulted in the highest performance on the validation set of different datasets, $k$ and $\omega$ values in range 2 to 8 and 1000 to 5000 were used respectively during the negative sampling step. In other words, the best performances on the validation sets in our empirical study were found by manual tuning of the hyperparameters. More information about the experimental trials can be found in Table \ref{tab:trial}, where the total number of trials for training each of the graph embedding models on each dataset is also indicated. 

\begin{table}[H]
    \centering
    \begin{tabular}{cccc}
    \hline
         SANS 
         & \multirow{2}{1cm}{\centering $k$-range}
         & \multirow{2}{1cm}{\centering $\omega$-set}
         & \multirow{2}{1cm}{\centering \#trials} \\
         Algorithm & & & \\
         \hline
         Uniform/ 
         & \multirow{2}{1cm}{\centering 2-8}   
         & \multirow{2}{0.9cm}{\centering N/A}   
         & \multirow{2}{0.9cm}{\centering 7} \\
         Self-Adv  & & &   \\   
         \hline
         Uniform RW/        
         & \multirow{2}{1cm}{\centering 2-5} 
         & \{1000, 1500,  
         & \multirow{2}{1cm}{\centering 32} \\
         Self-Adv RW & & \dots, 4500\} &    \\
         \hline
    \end{tabular}
    \caption{Hyperparameter combination sets and number of trials per model.}
    \label{tab:trial}
\end{table}

% ----------------------------------------------------------

% \begin{table*}[htbp!]
%     \centering
%     \begin{tabular}{cccc}
%     \hline
%          SANS Algorithm & $k$-range & $\omega$-set & \#trials \\
%          \hline
%          Uniform/Self-Adv & 2-8 & N/A & 7 \\
%          \hline
%         %  \multirow{2}{3cm}{\centering Uniform RW/Self-Adv RW }&  \multirow{2}{1cm}{\centering 2-5}
%         %  &1000, 1500,  & \multirow{2}{3cm}{\centering 27}\\
%         %  & & 2000, \dots ,4500, 5000 &  \\
%          Uniform RW/Self-Adv RW & 2-5 & \{1000, 1500, \dots , 4500\} & 32 \\
%         %  & &  &  \\
%          \hline
%     \end{tabular}
%     \caption{Hyperparameter Combination Sets and Number of Trials per Model.}
%     \label{tab:trial}
% \end{table*}

% % ---------------------------------------------------------- % The SANS hyperparameter tables

Additionally, Table \ref{tab:param} lists the hyperparameters with which different graph embedding models were trained to reach their optimal performance on the validation sets.

\begin{table}[H]
\centering
\begin{tabular}{cccc}
\hline
Hyper-          
& \multirow{2}{1.1cm}{\centering TransE}     
& \multirow{2}{1.4cm}{\centering DistMult}
& \multirow{2}{1.1cm}{\centering RotatE} \\ 
parameter & & &  \\
\hline
Embedding     
&  \multirow{2}{1.1cm}{\centering1024}     
&  \multirow{2}{1.4cm}{\centering1024}     
&  \multirow{2}{1.1cm}{\centering1024} \\ 
Dimension               &               &           &           \\
Batch Size              &      1000     &  2000     &  1000     \\ 
$\gamma$                &      9        &  200      &  9        \\  
Optimizer               &      Adam     &  Adam     & Adam      \\ 
$\alpha$                &      5E-05    &  1E-03    & 5E-05     \\  
\hline
\end{tabular}
\caption{Graph Embedding Models' Hyperparameters.}
\label{tab:param}
\end{table}
 % the embedding models' hyperparameters

% ----------------------------------------------------------

\begin{table*}[ht]
    \small
    \centering
    \begin{tabular}{cccccc}
    \hline
    \multirow{3}{2cm}{\centering Anchor Node} & 
    \multicolumn{5}{c}{Candidate Nodes} \\
    & \multirow{2}{3cm}{\centering Uniform} & \multicolumn{4}{c}{Uniform SANS}\\
    & & $k=2$ & $k=3$ & $k=4$ & $k=5$ \\
    \hline
    \multirow{5}{2cm}{\centering arachnoid} & diner & arachnida & biological & plectognathi & actinidia \\
    & refusal & biology & ostracoda & neritidae & bangiaceae \\
    & landscape & arthropod & subkingdom & amphibian\_family & barn\_spider \\
    & rise & wolf spider & placodermi & pelecaniformes & holarrhena \\
    & nurser & garden spider & scyphozoa & categorize & lucilia \\
    \hline
    \multirow{5}{2cm}{\centering empathy} & beach pea & sympathetic & sympathizer & cheerlessness & cheerfulness \\
    & sanvitalia & sympathy & expectation & ambition & pleasure \\
    & albinism & feeling & passion & have a bun in the oven & sympathize \\
    & micromeria & commiserate & pride & pleasure & enjoyment \\
    & banking industry & commiseration & state & attribute & stimulate \\
    \hline
    \multirow{5}{2cm}{\centering wheat} & lend & wild rice & fast food & Edirne & Jena \\
    & align & tabbouleh & salad & United States & seasoning \\
    & doodad & barley & mess & fixings & Washington \\
    & mismanage & Bulgur & stodge & form & pudding  \\
    & semiconductor device & buckwheat & meal & Iraqi Kurdistan & Bursa \\
    \hline
    \end{tabular}
    \caption{Example set of candidate nodes to form a negative triplet given an anchor node produced by uniform sampling (Uniform) and Uniform SANS. In this table, $k$ refers to the radius of the $k$-hop neighbourhood from which the candidate nodes are drawn by SANS.}
    \label{tab:neg_examples}
\end{table*}

% ----------------------------------------------------
% ----------------------------------------------------
% ----------------------------------------------------
% ----------------------------------------------------

 %negatives
\begin{table*}[!htbp]
    \centering
    \small
    \begin{tabular}{cccccccc}
          \hline
         \multirow{2}{1cm}{\centering Dataset} & \multirow{2}{2cm}{\centering Score Function} & 
         \multirow{2}{2cm}{\centering SANS Algorithm} &
         \multirow{2}{1cm}{\centering \emph{k}} & 
         \multicolumn{2}{c}{H@10} & \multicolumn{2}{c}{\centering MRR} \\ 
         & & & & Validation & Test & Validation & Test \\ \hline
         % ----------------------------------------------------
         \multirow{6}{1cm}{\centering FB15K-237} & 
         \multirow{2}{1cm}{\centering TransE} & Uniform &	3 & 48.55 &	48.35 & 0.3010 & 0.2962 \\
         & & Self-Adversarial &	3 &	52.51 &	52.03 &	0.3340 &	0.3265 \\
         & \multirow{2}{1cm}{\centering DistMult} & Uniform &	3 &	40.86 &	41.00 &	0.2599 &	0.2595 \\
         & & Self-Adversarial &	3 &	49.07 &	48.68 &	0.3131 &	0.3100 \\
         & \multirow{2}{1cm}{\centering RotatE} &	Uniform &	3 &	48.64	 & 48.22 &	0.3031 &	0.2985 \\
         & & Self-Adversarial &	5 &	53.72 &	53.12 &	0.3432 &	0.3358 \\
         \hline
         \multirow{6}{1cm}{\centering WN18}	& \multirow{2}{1cm}{\centering TransE} &	Uniform &	5	& 94.97 &	95.09 &	0.8237 &	0.8228 \\
         & & Self-Adversarial &	5 &	84.61 &	84.06 &	0.7165 &	0.7136 \\
         & \multirow{2}{1cm}{\centering DistMult} & Uniform &	3	& 93.07 & 93.19 &	 0.7507 &	0.7553 \\
         & & Self-Adversarial &	3 &	92.90 &	93.04 & 0.7534 &	0.7561 \\
         & \multirow{2}{1cm}{\centering RotatE} &	Uniform	 & 4 &	95.69 & 95.97 &	0.9492 & 0.9499 \\
         & & Self-Adversarial &	5 &	95.61	& 95.85 &	0.9489	& 0.9494 \\
         \hline
         \multirow{6}{1cm}{\centering WN18RR} & \multirow{2}{1cm}{\centering TransE} & Uniform &	4	& 50.89 &	51.15 &	0.2228 &	0.2254 \\
         & & Self-Adversarial &	8 &	52.46 &	53.21 &	0.2207 & 0.2249 \\
         & \multirow{2}{1cm}{\centering DistMult}	& Uniform &	6 & 44.73 & 44.74 &	0.4047 &	0.4025 \\
         & & Self-Adversarial &	8 &	39.01 &	38.70 &	0.3749 &	0.3684 \\
         & \multirow{2}{1cm}{\centering RotatE} &	Uniform &	4 &	55.78 &	55.76 &	0.4816 &	0.4769 \\
         & & Self-Adversarial &	8 &	56.76 &	57.12 &	0.4788 &	0.4745 \\
         \hline
    \end{tabular}
    \caption{The hyperparameter values associated with the best performance on the validation sets, used for obtaining the test results. The different variations of SANS in this table explicitly compute the \emph{k}-hop adjacency tensor.}
    \label{tab:ajdSANS}
\end{table*}

% ---------------------------------------------------------
% ---------------------------------------------------------
% ---------------------------------------------------------

\begin{table*}[!htbp]
    \centering
    \small
    \begin{tabular}{ccccccccc}
    \hline
         \multirow{2}{1cm}{\centering Dataset} & \multirow{2}{2cm}{\centering Score Function} & 
         \multirow{2}{2cm}{\centering SANS Algorithm} &
         \multirow{2}{1cm}{\centering \emph{k}} & 
         \multirow{2}{1cm}{\centering $\omega$} &
         \multicolumn{2}{c}{H@10} & \multicolumn{2}{c}{\centering MRR} \\ 
         & & & & & Validation & Test & Validation & Test \\ \hline
         % ---------------------------------------------------- %FB15K-237
         \multirow{6}{1cm}{\centering FB15K-237} &	
         \multirow{2}{2cm}{\centering TransE}	
         & Uniform &	5 &	4000 &	49.12 &	48.50 &	0.3023 &	0.2981 \\
         & & Self-Adversarial &	4 &	4000&	50.59 &	50.04 &	0.3129 &	0.3060  \\
         &	\multirow{2}{2cm}{\centering DistMult} &	Uniform	& 4	& 3000 & 41.66 &	41.46 &	0.2628 &	0.2621 \\
         & & Self-Adversarial &	5	& 3000 & 48.67  & 48.17	& 0.3142	 & 0.3071 \\ 
         & \multirow{2}{2cm}{\centering RotatE}  &	Uniform &	2 &	4000 &	49.05 &	48.47 &	0.3034 &	0.3003 \\
         & & Self-Adversarial &	2	& 4000	& 51.41 &	51.07 &	0.3205 &	0.3161 \\
         \hline 
         \multirow{6}{1cm}{\centering WN18} &	
         \multirow{2}{2cm}{\centering TransE} 
         & Uniform &	2	& 1000 &	95.23 &	95.22 &	0.8194 &	0.8195 \\
         & & Self-Adversarial &	3 &	4000 &	88.65 & 88.51 &	0.7480 &	0.7429 \\
         & \multirow{2}{2cm}{\centering DistMult}	& Uniform &	2 &	1000 &	89.38 & 89.80 &	0.6205 &	0.6235 \\
         & & Self-Adversarial &	2 &	1000 &	90.55 & 91.08 &	0.6601 &	0.6634 \\
         &  \multirow{2}{2cm}{\centering RotatE}  &	Uniform & 2 & 3000 & 95.92 & 96.07 &	0.9492 & 0.9489 \\
         & & Self-Adversarial &	2 &	4500 &	95.83 & 96.09 &	0.9493 &	0.9496 \\
         \hline 
         \multirow{6}{2cm}{\centering WN18RR} 	&  \multirow{2}{2cm}{\centering TransE}
         &	Uniform &	2 &	1000 &	52.67&	53.41&	0.2282 &	0.2317 \\
         & & Self-Adversarial &	5 &	2000 &	53.05 &	53.81 &	0.2229 &	0.2273 \\
         & \multirow{2}{2cm}{\centering DistMult} & 	Uniform &	2 &	3000 &	49.01 &	49.09 &	0.4111 &	0.4071 \\
         & & Self-Adversarial &	4 &	1000 &	43.70 &	42.74	& 0.3883 & 0.3836 \\
         & \multirow{2}{2cm}{\centering RotatE} 
         &	Uniform	& 2	& 1000	& 57.20	& 57.12 &	0.4860 &	0.4796 \\
         & & Self-Adversarial &	2 &	1000 &	57.09 &	56.94 &	0.4882 &	0.4805 \\
        \hline
    \end{tabular}
    \caption{The hyperparameter values associated with the best performance on the validation sets, used for obtaining the test results. The different variations of SANS in this table  approximate the \emph{k}-hop adjacency tensor by random walks (RW-SANS) using Alg. \ref{alg:rw}.}
    \label{tab:rwSANS}
\end{table*}
% ---------------------------------------------------------
% % ------------------------------------------------------------------------
% COMPLEXITY TABLE START
\begin{table*}[ht]
    \centering
    \begin{tabular}{cccc}
    \hline
         \multirow{2}{3cm}{\centering Negative Sampling Algorithm} & Preprocessing  & Runtime  & Space \\
         & Complexity & Complexity & Complexity
         \\
         \hline
         Uniform \cite{bordes2013translating} & $O(1)$ & $O(bn)$ & $O(1)$ \\
        %  IGAN \cite{yu2014improving} & $O(t)$ & $O(bnt + bt)$ & $O(t)$ \\
         KBGAN \cite{cai2017kbgan} & $O(t)$ & $O(bn + bd + bt)$ & $O(t)$ \\
         NSCaching \cite{zhang2019nscaching} & $O(1)$ & $O(bn + be)$ & $O(c|R||V|)$ \\
         Self-Adv. \cite{sun2019rotate} & $O(|E|)$ & $O(bn + bd)$ & $O(|E|)$ \\
         Uniform SANS (ours) & $O(|V|^3\log k)$ & $O(bn)$ & $O(|V|^2)$ \\
         Self-Adv. SANS (ours) & $O(|V|^3\log k)$ & $O(bn + bd)$ & $O(|V|^2)$ \\
         Uniform RW-SANS (ours) & $O(rk|V|)$ & $O(bn)$ & $O(r|V|)$ \\
         Self-Adv. RW-SANS (ours) & $O(rk|V|)$ & $O(bn + bd)$ & $O(r|V|)$ \\
         \hline
    \end{tabular}
    \caption{Comparison of different negative sampling algorithms in terms of preprocessing, runtime, and space complexities given batch size $b$, negative sample size $n$, cache size $c$, cache extension size $e$, node set $V$, edge set $E$, relation set $R$, embedding dimension $d$, hops count $k$, random walks count $r$, and GAN parameters count $t$.}
    \label{tab:times}
\end{table*}
% COMPLEXITY TABLE END
% % ------------------------------------------------------------------------
 % complexity table

\subsection{Preprocessing}

Building the \emph{k}-hop neighborhood of the nodes within the KG can be regarded as the preprocessing step, essential to implementing SANS. In this paper, we propose two techniques for doing so, which are:
\begin{enumerate}
    \item explicit computation of the \emph{k}-hop neighborhood by manipulating Eqn. \ref{eqn:khop} while accounting for different relation types and,
    \item approximation of the \emph{k}-hop neighborhood using random walks, as detailed in Alg. \ref{alg:rw}.
\end{enumerate}

% ----------------------------------------------------------

\subsection{Infrastructure Settings}

The experiments in our study were carried on a server with one NVIDIA V100 GPU, 10 CPU cores, and 46GB RAM.

% ----------------------------------------------------------

\section{Experimental Results}

\subsection{Qualitative Assessment of Negative Samples}
\label{qualitative_sans}

In this section, we assess the semantic meaningfulness of negative samples produced by Uniform SANS and those produced by uniform sampling using the WN18RR dataset. As presented by the examples given in Table \ref{tab:neg_examples}, Uniform SANS results in negative examples that are harder to distinguish semantically compared to uniform sampling. We also notice the semantic meaningfulness of SANS-based negatives decline as we increase the size of the $k$-hop neighbourhood. This observation is indeed expected since as the neighbourhood increases in size (i.e. $k \to \infty$) Uniform SANS will become analogous to uniform sampling.

% ----------------------------------------------------------

\subsection{SOTA Algorithms}

Results for the Uniform and Self-Adversarial algorithms in Table \ref{tab:unisota_comp} and Table \ref{tab:comp_adv} respectively were achieved by re-running the code provided by ~\cite{sun2019rotate} using the hyperparameters they reported for the best performance on the validation set of different datasets. Additionally, the results for KBGAN and NSCaching in Table \ref{tab:unisota_comp} are the \emph{scratch} results directly taken from \cite{zhang2019nscaching}.

% ----------------------------------------------------------

\subsection{SANS Algorithms}

Table \ref{tab:ajdSANS} and Table \ref{tab:rwSANS} report the performance of the graph embedding models fused with our negative sampling techniques on the validation and test sets with respect to the evaluation metrics. Additionally, they list the hyperparameter values corresponding to Uniform/Self-Adv. SANS and Uniform/Self-Adv. RW-SANS that resulted in the best performance on the validation sets. Based on our outcomes, we hypothesize that the usage of random walks in approximating the \emph{k}-hop neighborhood implicitly results in the removal of nodes with the least number of walks to the center node---i.e. outlier nodes.

% ----------------------------------------------------------

\section{Computational Complexity}

Table \ref{tab:times} is representative of the time and space complexities of different negative sampling approaches including SANS.

\end{document}